# Heart Disease Detection using Vision-Based Transformer Models from ECG Images

Zeynep Hilal Kilimci*, Mustafa Yalcin, Ayhan Kucukmanisa and Amit Kumar Mishra

*Abstract*—Heart disease, also known as cardiovascular disease, is a prevalent and critical medical condition characterized by the impairment of the heart and blood vessels, leading to various complications such as coronary artery disease, heart failure, and myocardial infarction. The timely and accurate detection of heart disease is of paramount importance in clinical practice. Early identification of individuals at risk enables proactive interventions, preventive measures, and personalized treatment strategies to mitigate the progression of the disease and reduce adverse outcomes. In recent years, the field of heart disease detection has witnessed notable advancements due to the integration of sophisticated technologies and computational approaches. These include machine learning algorithms, data mining techniques, and predictive modeling frameworks that leverage vast amounts of clinical and physiological data to improve diagnostic accuracy and risk stratification. In this work, we propose to detect heart disease from ECG images using cutting-edge technologies, namely vision transformer models. These models are Google-Vit, Microsoft-Beit, and Swin-Tiny. To the best of our knowledge, this is the initial endeavor concentrating on the detection of heart diseases through image-based ECG data by employing cuttingedge technologies namely, transformer models. To demonstrate the contribution of the proposed framework, the performance of vision transformer models are compared with state-of-the-art studies. Experiment results show that the proposed framework exhibits remarkable classification results.

## I. INTRODUCTION

Cardiovascular disease , commonly referred to as heart disease, is a prevalent and critical medical condition characterized by the dysfunction of the heart and blood vessels, leading to various complications such as coronary artery disease, heart failure, and myocardial infarction. Its global impact on mortality and morbidity necessitates a comprehensive understanding and effective detection strategies. The accurate and timely detection of heart disease assumes paramount significance within the realm of clinical practice. Early identification of individuals at risk enables proactive interventions, preventive measures, and personalized treatment modalities to effectively manage disease progression and mitigate adverse outcomes. Detection methodologies play a pivotal role in identifying individuals with preclinical or asymptomatic stages of heart disease, thus facilitating early intervention and the implementation of tailored lifestyle modifications. Significant advancements have been witnessed in recent years in the field of heart disease detection, driven by the integration of sophisticated technologies and computational approaches. These include the utilization of machine learning algorithms, data mining techniques, and predictive modeling frameworks, capitalizing on extensive clinical and physiological data to enhance diagnostic accuracy and risk stratification. The application of these advanced techniques holds great promise in augmenting the sensitivity and specificity of diagnostic tests, enabling more precise and reliable identification of individuals at risk. The precise detection of heart disease also enables the optimal allocation of healthcare resources. By identifying individuals at a high risk, healthcare practitioners can prioritize interventions and allocate resources efficiently, ensuring timely and appropriate care for patients. Moreover, early detection can contribute to a reduction in healthcare costs associated with advanced disease stages and their subsequent complications. Furthermore, the detection of heart disease plays a pivotal role in advancing our comprehension of the underlying mechanisms and etiology of the condition. Through comprehensive analysis of extensive datasets and the identification of pertinent biomarkers, researchers can unravel the intricate interplay between genetic, environmental, and lifestyle factors, thereby paving the way for targeted interventions and the development of personalized medicine approaches. In the recent past, deep learning is a prominent branch of machine learning that focuses on training artificial neural networks with multiple layers to extract abstract representations from intricate data. It has garnered considerable attention and achieved noteworthy success across diverse domains, including computer vision, natural language processing, and speech recognition. Transformer models, specifically the Transformer architecture, have emerged as a groundbreaking paradigm in deep learning for tasks involving sequences, such as natural language understanding and generation. The primary innovation of Transformer models lies in the selfattention mechanism, which enables the capture of global dependencies and interrelationships among different elements in the input sequence. This mechanism facilitates efficient parallelization during both training and inference, endowing Transformers with scalability and effectiveness in processing lengthy sequences. Deep learning and Transformer models offer formidable capabilities for classification tasks. They possess the capacity to autonomously learn discriminative features from raw input data, eliminating the necessity for manual feature engineering. Notably, convolutional neural networks (CNNs) and recurrent neural networks (RNNs), two prevalent deep learning models, have exhibited outstanding performance in image and text classification tasks, respectively. The applications of deep learning and Transformer models in classification tasks span a broad spectrum. They have been effectively employed in sentiment analysis, document classification, object recognition, spam detection, medical diagnosis, and various other domains. Their capacity to discern intricate patterns and extract meaningful representations from extensive datasets endows them with substantial value for addressing classification problems across diverse disciplinary contexts. In this work, our objective is to develop an advanced methodology for the detection of heart disease based on electrocardiogram (ECG) images, employing state-of-the-art technologies, specifically vision transformer models. The vision transformer models considered for evaluation in this research encompass Google-Vit, Microsoft-Beit, and Swin-Tiny architectures. To highlight the novelty and effectiveness of our proposed framework, we conduct a comprehensive comparative analysis by benchmarking the performance of the vision transformer models against well-established deep learning methodologies, including convolutional neural networks (CNN) and Residual Networks (Res-Net), as well as the most recent state-of-theart studies. The experimental outcomes affirm the exceptional classification performance achieved by Swin-Tiny model with 96.63% of accuracy, surpassing the performance exhibited by existing literature studies. The main contributions of this study are as follows:

• Performing vision-based transformer frameworks for the purpose of detecting heart disease.

• Comparing the classification performance of vision-based transformer approaches, namely VIT, SWIN, and BEIT.

• Providing superior classification success by conducting an extensive comparison of the transformer-based framework with the literature studies.

The rest of the paper is designated as: Section 3 presents a brief summary of literature studies on ECG classification. Section 4 introduces deep learning and transformer models employed in the study. Methodology including data collection and proposed framework steps is presented in Section 5. Section 6 details experiment results. The paper concludes with conclusions and discussions in Section 7.

## II. Related Work

In this section, a concise overview is provided regarding the identification of diseases from ECG image data through the utilization of diverse methods.

In [1], the ECG samples of the subjects are taken into account as the necessary inputs for the HD detection model. Numerous useful articles for classifying HD using various machine learning (ML) and deep learning (DL) models have been reported in the recent past. It is noted that, when dealing with imbalanced HD data, the detection accuracy is reduced. With the aim of achieving improved HD detection, appropriate DL and ML models are identified in this study, and the requisite classification models are developed and evaluated. The

Generative Adversarial Network (GAN) model is selected to address imbalanced data by generating and utilizing additional synthetic data for detection purposes. Furthermore, an ensemble model comprising long short-term memory (LSTM) and GAN is formulated in this paper, demonstrating superior performance compared to individual DL models employed in this study. The simulation results, based on the standard MIT-BIH dataset, reveal that the proposed GAN-LSTM model yields the highest accuracy, F1-score, and area under the curve (AUC) at 0.992, 0.987, and 0.984, respectively, in comparison to other models. Likewise, for the PTB-ECG dataset, the GAN-LSTM model surpasses all other models with accuracy, F1-score, and AUC scores of 0.994, 0.993, and 0.995, respectively. It is observed that among the five models under investigation, the GAN model exhibits the best performance, whereas the detection potential of the NB model is the lowest.

In [2] A cardiac disorder detection system from 12-leadbased ECG images is proposed. Various ECG equipment is utilized by healthcare institutes, and results are presented in nonuniform formats of ECG images. A generalized methodology to process all formats of ECG is proposed by the study. Cardiovascular disease detection is carried out using a Single Shoot Detection (SSD) MobileNet v2-based Deep Neural Network architecture. The four major cardiac abnormalities (i.e., myocardial infarction, abnormal heartbeat, previous history of MI, and the normal class) are focused on, and accuracy results are calculated at 98%. The dataset, comprising 11,148 standard 12-lead-based ECG images used, is manually collected from healthcare institutes and annotated by domain experts. High accuracy results for differentiating and detecting the four major cardiac abnormalities are achieved. The proposed system's accuracy result is manually verified by several cardiologists, and it is recommended that the system can be used to screen for a cardiac disorder.

In [3], a two-stage multiclass algorithm is proposed. In the first stage, ECG segmentation is performed based on Convolutional Bidirectional Long Short-Term Memory neural networks with an attention mechanism. In the second stage, a time adaptive Convolutional Neural network is applied to ECG beats that have been extracted from the first stage over several time intervals. ECG beats are transformed into 2D images using the Short-Time Fourier Transform to automatically distinguish normal ECG from cardiac adverse events, such as arrhythmia and congestive heart failure, and to predict sudden cardiac death. Model accuracy is compared across different time scales. The data used to train and test the models is extracted from MIT/BIH-PhysioNet databases. With the use of 4-minute ECG data, congestive heart failure events can be automatically detected with an accuracy of 100%, arrhythmia events with 97.9%, and sudden cardiac deaths with 100%.

The objective of the study [4] is to employ a deep-learning approach utilizing image classification for the detection of heart disease. Image recognition is currently dominated by a deep convolutional neural network (DCNN) as the preferred classification technique. The public UCI heart disease dataset, which includes 1050 patients and 14 attributes, is being used to evaluate the proposed model. By collecting a set of features directly extractable from the heart disease dataset, this feature vector is considered as the input for a DCNN to determine whether an instance is classified as belonging to either the healthy or cardiac disease category. To evaluate the performance of the method, various performance metrics, including accuracy, precision, recall, and the F1 measure, are applied, and a validation accuracy of 91.7% is achieved by proposed model.

In [5], a method for diagnosing two types of heart arrhythmia using the ECG record as an image is proposed. To assess the performance of the system, five feature extraction methods widely recognized in the literature are employed, and five different classifiers are tested. Heart disorders are successfully identified with an accuracy exceeding 96.00% by employing a vanilla neural network, Multilayer Perceptron (MLP), and Local Binary Patterns (LBP) on ECG images. Promising results have been demonstrated from a medical perspective through this investigation.

In [6], a methodology for the extraction of multiple feature vectors from ultrasound images of carotid arteries (CAs) and the assessment of heart rate variability (HRV) in electrocardiogram signals is introduced. Additionally, a suitable and reliable prediction model for the diagnosis of cardiovascular disease (CVD) is presented. The creation of multiple feature vectors involves the extraction of a candidate feature vector through image processing and the measurement of carotid intimamedia (IMT) thickness. In addition, linear and/or nonlinear feature vectors are derived from HRV, a primary indicator of cardiac disorders. The significance of these multiple feature vectors is assessed using various machine learning methods, including Neural Networks, Support Vector Machine (SVM), Classification based on Multiple Association Rule (CMAR), Decision tree induction, and Bayesian classifier. The results demonstrate that multiple feature vectors extracted from both CAs and HRV (CA+HRV) yield higher accuracy compared to separating the feature vectors of CAs and HRV. Moreover, the SVM and CMAR exhibit approximately 89.51% and 89.46%, respectively, in terms of diagnostic accuracy when using the ultimately selected multiple feature vectors for diagnosis or prediction.

In [7], a hybrid diagnostic tool is developed, which integrates various machine learning techniques. These techniques are capable of analyzing clinical histories and electrocardiogram signal images, determining whether the patient has ischemic heart disease with an accuracy of up to 97.01%. Collaboration with medical experts and the utilization of a database containing clinical data for around 1020 patients and their diagnoses are essential components of this project. Additionally, a picture database containing 92 electrocardiogram signal images is also employed in this study for the analysis by the Artificial Neural Network.

In [8], a new Deep Learning (DL) approach is being introduced for the automated identification of Congestive Heart Failure (CHF) and Arrhythmia (ARR) with high accuracy and minimal computational demands. For the first time, a novel ECG diagnosis algorithm is presented, which combines the Convolutional Neural Network (CNN) with the ConstantQ Non-Stationary Gabor Transform (CQ-NSGT). The CQNSGT algorithm is employed to transform the 1-D ECG signal into a 2-D time-frequency representation, which is then supplied to a pre-trained CNN model known as AlexNet. The features extracted through the AlexNet architecture are used as pertinent features to be distinguished by a Multi-Layer Perceptron (MLP) technique in three distinct cases: CHF, ARR, and Normal Sinus Rhythm (NSR). The performance of the proposed CNN with CQ-NSGT is compared to CNN with Continuous Wavelet Transform (CWT), revealing the effectiveness of the CQ-NSGT algorithm. The approach is evaluated with real ECG records, and the experimental results demonstrate the superior performance of the proposed method in terms of accuracy (98.82%), sensitivity (98.87%), specificity (99.21%), and precision (99.20%).

The study [9] is aimed at developing algorithmic models for the analysis of ECG images to predict cardiovascular diseases. The primary impact of this work is the saving of lives and the enhancement of medical care at reduced costs. With the increasing costs of healthcare and health insurance worldwide, the direct result of this work is the preservation of lives and the improvement of medical care. Numerous experiments have been conducted to optimize deep learning parameters. The same validation accuracy value of approximately 0.95 is found for both the MobileNetV2 and VGG16 algorithms. Upon implementation on Raspberry Pi, a slight decrease in accuracy is observed (0.94 and 0.90 for MobileNetV2 and VGG16 algorithms, respectively). Consequently, the primary objective of this research is to enhance real-time monitoring using costeffective and accessible smart mobile tools, including mobile phones, smartwatches, connected T-shirts, and similar devices.

In [10], a machine learning-based approach for predicting heart disease is introduced, utilizing a heart disease dataset. Furthermore, the system proposed here can easily differentiate and classify individuals with heart disease from those who are healthy. The Cleveland heart disease dataset is utilized in this study, incorporating ECG images for the creation of a hybrid model. Essential features are then extracted using the Genetic Algorithm and PSO algorithm. Subsequently, a neural network algorithm is employed to construct a prediction model. This prediction model is then applied to test data to compute metrics such as prediction accuracy. As a result, the machine learningbased approach proposed in this study, as part of the decision support process, assists medical practitioners in effectively diagnosing heart patients.

## III. PROPOSED FRAMEWORK

The proposed framework encompasses the analysis of ECG images using three distinct deep vision transformer approaches. In conducting this analysis, we initially opted for a well-established ECG dataset [11]. The objective of this study is to provide a solution for the detection of heart disease in patients through the utilization of different vision transformer methods applied to ECG images. The block diagram of proposed framework is shown in Figure 1.

### A. Dataset

ECG images dataset [11] comprises 1937 individual patient records. Patient data in this dataset is obtained from ECG Device 'EDAN SERIES-3' installed in Cardiac Care and Isolation Units within various healthcare facilities throughout Pakistan. Example image from ECG dataset is presented in Figure 2. The ECG image data, painstakingly collected, underwent meticulous manual review by medical experts utilizing the Telehealth ECG diagnostic system. This review

### B. Preprocessing

An ECG report conventionally includes both alphanumeric values and waveform graphs, as depicted in Figure 2. The upper section of the ECG report often comprises a list of alphanumeric values, which typically does not impact the detection process. The waveform charts, commonly occupying the central and lower sections of the ECG, represent timeseries graphs derived from sensor data. In this study, the critical focus lies on processing these waveform data. To facilitate this, a region of interest (ROI) is identified within the image, specifically containing the 12-lead waveforms, and subsequently this area is cropped. The outcome of this cropping operation applied to Figure 2 is presented in Figure 3.

In the detection process, the crucial information lies in the waveform's structure. Therefore, the presence of 3dimensional color information in the image adds unnecessary processing overhead. Moreover, there are

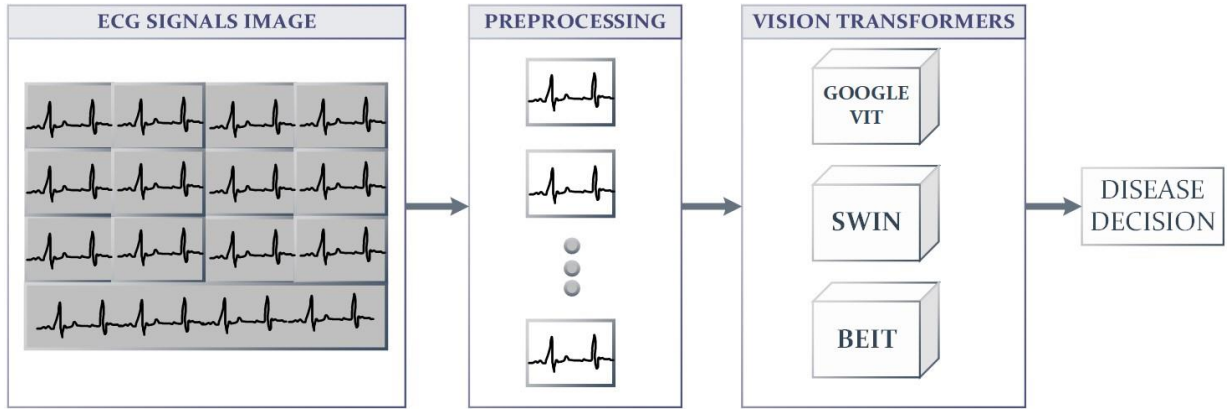

Fig. 1: The block diagram of proposed framework

process, conducted under the supervision of seasoned medical .

professionals proficient in ECG interpretation, extended over several months. It encompassed the assessment of ECG images across five distinct categories: COVID-19, Abnormal Heartbeat, Myocardial Infarction (MI), Previous History of MI, and Normal Individuals.

The compiled dataset features 12-lead ECG images, rendering it a valuable resource for Data Scientists, IT Professionals, and Medical Research Institutes. This dataset is primarily designed to support research focused on COVID-19, Arrhythmia, and a wide range of other cardiovascular conditions. In our work, we focus on heart disease conditions. The diseases categories and the corresponding number of images (report x 12-lead image) in the dataset are as follows:

- Myocardial Infarction Patients (240x12=2880)
- Patient that have abnormal heartbeat (233x12=2796)
- Patient that have History of MI (172x12=2064)
- Normal (172x12=2064)

unwanted dot-shaped structures in the background of the image. To address these two undesirable aspects, the image is binarized using global thresholding method. The threshold value of 40 is employed, as it yielded the most favorable results. Binarization process results on Figure 3 is shown in Figure 4.

In the final step, the resulting binarized image, which includes all 12 lead images, is divided into separate subimages, with each subimage corresponding to a single lead image. This process is visually depicted in Figure 5.

### C. Proposed Methods

Vision transformers represent a potent approach that can address a multitude of complex challenges across various domains. In the context of our study, we employed three distinct vision transformers: Google-Vit [12], Swin [13], and BEiT [14], to accomplish heart disease detection.

*1) Google-Vit:* The Google Vision Transformer Model [12] is an advanced deep learning architecture designed specifically for visual recognition tasks. It represents a groundbreaking approach that amalgamates the prowess of transformer models, originally devised for natural language processing, with the domain of computer vision. The operational logic of the Vision Transformer Model revolves around a selfattention mechanism, enabling the model to capture global interdependencies and contextual information within an image. Departing from conventional Convolutional Neural Networks (CNNs) reliant on manually engineered features, the Vision Transformer Model harnesses the attention mechanism to learn pertinent visual representations directly from the raw image data.

Fundamentally, the Vision Transformer Model partitions an input image into smaller patches, treating them as sequential tokens. These patches are subsequently fed into a transformer encoder comprising multiple layers of self-attention and feedforward neural networks. The self-attention mechanism empowers the model to selectively focus on distinct regions of the image and glean their intricate relationships, while the feed-forward networks process the attended information to generate meaningful visual embeddings. During the training phase, the Vision Transformer Model is trained utilizing vast image datasets, such as ImageNet, replete with copious labeled images. Through supervised learning, the model acquires proficiency in predicting the correct class labels for the images, optimizing its parameters via techniques such as gradient descent and backpropagation.

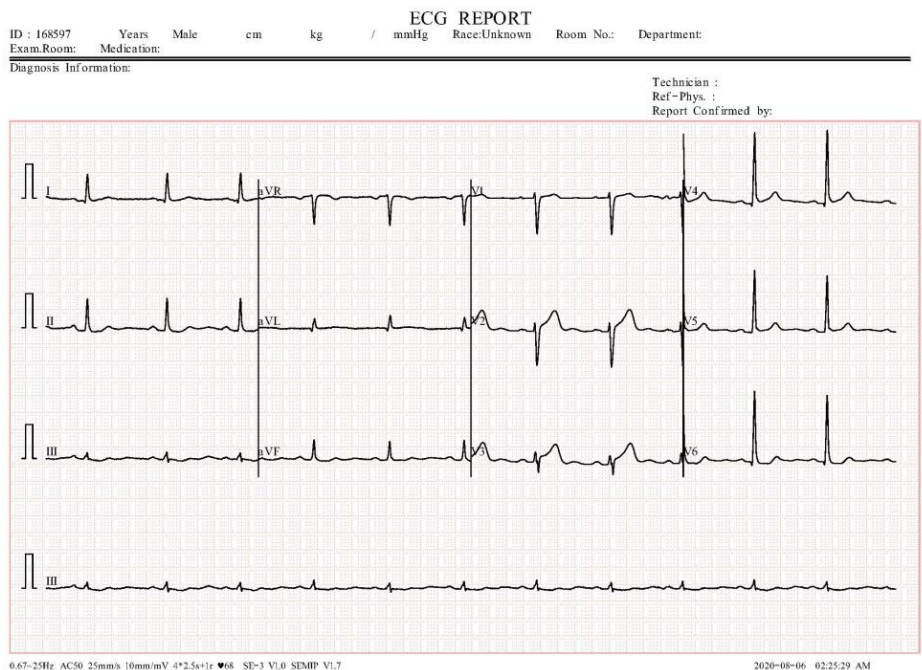

Fig. 2: The example ECG image from [11]

.

Fig. 4: Visualization of binarization process

.

The Vision Transformer Model has showcased remarkable performance across an array of computer vision tasks, encompassing image classification, object detection, semantic segmentation, and image generation. It has surpassed

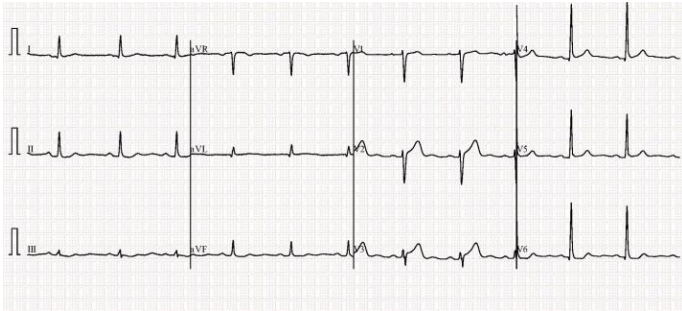

Fig. 3: Visualization of cropping process
.

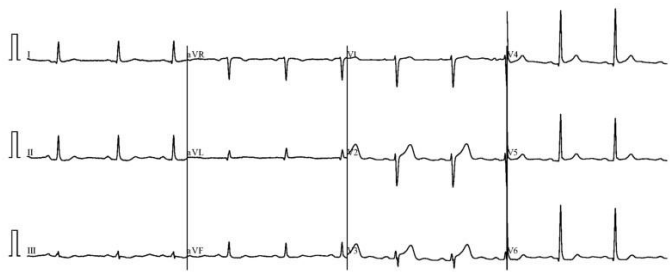

conventional CNN-based methodologies, attaining state-of-the-art outcomes .

on renowned benchmarks such as ImageNet. Its aptitude for capturing long-range dependencies and harnessing the selfattention mechanism renders it particularly efficacious for tasks involving intricate and large-scale visual data. Furthermore, the Vision Transformer Model offers salient advantages, including scalability and interpretability. It can effectively handle images of diverse sizes and adapt to varying levels of intricacy. Moreover, the attention mechanism empowers researchers and practitioners to scrutinize and explicate the model's decision-making process, unraveling insights into the factors shaping its predictions.

To summarize, the Google Vision Transformer Model epitomizes a paradigm shift in computer vision, harnessing the formidable capabilities of transformer models for visual recognition tasks. Through its self-attention mechanism and ability to capture global interdependencies, it has spearheaded breakthroughs across diverse application domains. By leveraging expansive datasets and advanced training methodologies, the Vision Transformer Model has revolutionized the frontiers of computer vision, unveiling new prospects for research and practical implementations.

*2) Swin:* The Swin [13], also known as the Smaller Variant Transformer Model, is an advanced deep learning architecture that has garnered substantial acclaim within the computer vision community. It represents a condensed variant of the Transformer model, purposefully crafted for visual recognition tasks. The operational logic of the Swin-Tiny model builds upon the foundational tenets of the Transformer architecture. It embraces a hierarchical approach to process visual data,

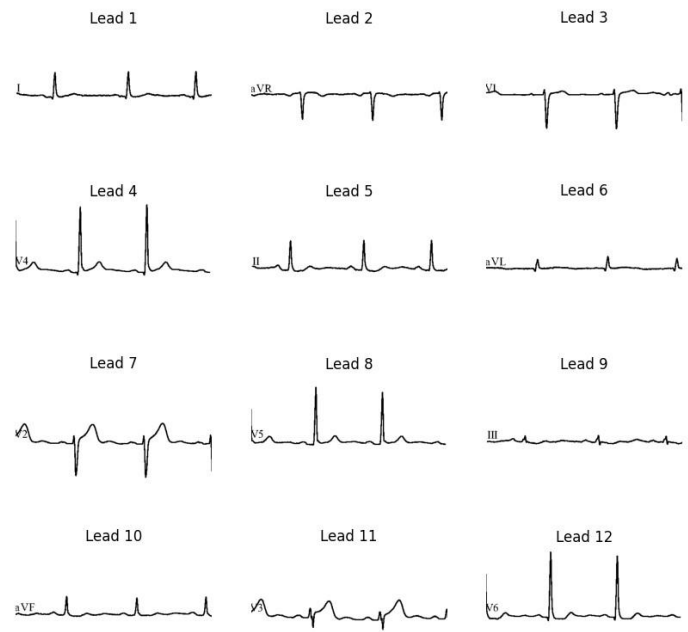

Fig. 5: Visualization of the process for separating lead images. fragmenting it into smaller patches akin to the Vision Transformer (ViT) model. Nonetheless, distinct from ViT, Swin-Tiny adopts a shift-based methodology to preserve computational efficiency while effectively capturing extensive dependencies.

The crux of the Swin-Tiny model revolves around the concept of shifted windows. Instead of concurrently processing all patches, it employs a partitioning strategy that incorporates shifts, resulting in non-overlapping windows. This strategy empowers the model to adeptly assimilate local and global contextual information. The Swin-Tiny model comprises a stacked arrangement of hierarchical transformer layers, orchestrated in a cascaded manner. Each layer encompasses two sub-layers: a shifted window self-attention mechanism and a feed-forward neural network. The shifted window selfattention mechanism operates at the patch level, enabling the model to selectively attend to distinct regions of the input image. This attention mechanism effectively captures contextual relationships between patches, fostering robust feature extraction. Throughout the training process, the Swin-Tiny model is typically trained on extensive image datasets, such as ImageNet, teeming with meticulously annotated labels. By means of supervised learning, the model acquires proficiency in accurately predicting labels for input images, optimizing its parameters through gradient descent and backpropagation techniques.

The Swin-Tiny model has evinced remarkable performance across a spectrum of computer vision tasks, encompassing image classification, object detection, semantic segmentation, and image generation. It rivals larger, computationally demanding models, endowing it with pronounced appeal in resourceconstrained scenarios. Moreover, the Swin-Tiny model has demonstrated its efficacy in handling diverse visual data, encompassing images of varying sizes and aspect ratios. Its capacity to capture extensive dependencies and leverage the shifted window self-attention mechanism contribute to its prowess in comprehending intricate visual patterns and structures.

In essence, the Swin-Tiny model represents a condensed rendition of the Transformer architecture, meticulously tailored for computer vision tasks. Its hierarchical design, shiftbased strategy, and self-attention mechanism facilitate the assimilation of local and global contextual information with remarkable efficiency. With its commendable performance across multifarious application domains and computational efficacy, the Swin-Tiny model holds significant promise for realworld applications, particularly those constrained by limited resources.

*3) BEiT:* The BEiT [14], which is based on the Vision Transformer (ViT) and operates as a transformer encoder akin to BERT, distinguishes itself from the original ViT by undergoing self-supervised pre-training on a substantial collection of images from ImageNet-21k at a resolution of 224x224 pixels. The pre-training objective involves predicting visual tokens derived from OpenAI's DALL-E's VQ-VAE encoder, employing masked patches. Following this pre-training phase, the model undergoes supervised fine-tuning on ImageNet (ILSVRC2012), an extensive dataset encompassing one million images and one thousand classes, also at a resolution of 224x224. To process images, the BEiT model treats them as a sequence of fixed-size patches (16x16 resolution) and linearly embeds them. Unlike the original ViT models, BEiT models utilize relative position embeddings similar to T5, as opposed to absolute position embeddings. Furthermore, image classification is achieved by mean-pooling the final hidden states of the patches, rather than applying a linear layer to the final hidden state of the [CLS] token. The addition of the [CLS] token serves to represent the entire image and can be employed for classification purposes.

Through pre-training, the model acquires an internal representation of images, which can then be utilized to extract features beneficial for downstream tasks. For instance, if a dataset of labeled images is available, a standard classifier can be trained by appending a linear layer atop the pretrained encoder. Typically, a linear layer is applied to the [CLS] token, as its last hidden state can be viewed as a representation of the entire image. Alternatively, the final hidden states of the patch embeddings can be mean-pooled, and a linear layer can be added on top of that representation. The BEiT model undergoes self-supervised pre-training on ImageNet-21k, encompassing 14 million images and 21,841 classes, at a resolution of 224x224, followed by fine-tuning on ImageNet

2012, comprising one million images and one thousand classes, at the same resolution. It was introduced in the paper titled "BEIT: BERT Pre-Training of Image Transformers" authored by Hangbo Bao, Li Dong, and Furu Wei and initially released in a specific repository.

## IV. Experimental Results

In this paper, we have opted to focus on four distinct types of clinical ECG waveform images from cardiac patients' records. These images are used as input for vision transformer methods to detect diseases. In the training and testing phases TABLE I: Training parameters

| Parameters | Google-Vit | Swin | BEiT |
|---|---|---|---|
| Epoch | 30 | 35 | 25 |
| Batch | 64 | 80 | 64 |
| Learning Rate | 9e-6 | 4e-5 | 6e-5 |
| Warmup Ratio | 0.1 | 0.1 | 0.08 |
| Optimizer | AdamW | AdamW | AdamW |

of vision transformer methods, a total of 817 images are utilized. Following the preprocessing steps, where each image is divided into 12 lead images, the total count expanded to 9,804. In the experiments, 5-fold cross-validation technique is applied to assess the performance of proposed methods. The reason for employing the K-fold strategy is to avoid potential bias that could arise from a simple 80% to 20% data split. This approach ensures a more equitable comparison while evaluating the model's generalization capability. We select commonly used metrics such as Accuracy, Precision, Recall, and F1-score to assess the performance of proposed methods. By considering all these metrics, we aim to provide a well-rounded assessment of our models' performance, taking into account factors like true positives, true negatives, false positives, and false negatives. This approach allows us to gauge not only the model's overall accuracy but also its ability to correctly classify positive and negative instances, which is especially important in medical diagnosis and disease detection tasks. The training parameters of the proposed methods are given in Table I. In our quest to determine the optimal configuration for the model, we systematically adjusted the model's parameter values and conducted training. To identify these optimal parameters, we employ the grid search technique. The training loss curves of proposed methods have been visualized and are presented in Figure 6. When the training curves are examined, there is not a significant gap between the losses, indicating that the model is not overfitting to the training data.

Table II provides a comparison of the proposed vision transformer models, showcasing their performance across precision, recall f1-score and accuracy metrics. As seen from table, all three vision transformer models exhibit strong performance in the ECG disease detection task. These results indicate that these models are well-suited for the detection of diseases in ECG images, providing a good balance between precision, recall, and overall accuracy. BEiT outperforms the other models across all metrics. This positions it as a highly promising candidate for disease detection in ECG images. Its outstanding precision, recall, F1-score, and accuracy make it a compelling choice for healthcare applications, demonstrating its potential to provide accurate and reliable diagnoses in this domain.

The proposed BEiT method is compared with recently proposed methods in Table 3. These methods employ the same dataset as proposed method. The result values for the methods compared in the table are sourced directly from their original papers. These benchmark methods are evaluated under different test setups, including cross-validation and holdout methods. To ensure a comprehensive assessment, our proposed method is also subjected to two different evaluation scenarios: a 5-fold cross-validation and a holdout split of 80/0/20 (train, TABLE II: Comparison of proposed vision transformers mod-els

| Models | Precision | Recall | F1-Score | Accuracy |
|--------|-----------|--------|----------|----------|
| Google-Vit | 0.943 | 0.943 | 0.942 | 0.943 |
| Swin | 0.955 | 0.955 | 0.954 | 0.955 |
| BEiT | 0.959 | 0.959 | 0.959 | 0.959 |

validation, test ratio). It is evident that our proposed method consistently outperforms the benchmark models across all metrics, except for Recall when compared to [15]. The lower recall of our proposed method in comparison to [15] could be attributed to differences in the dataset's distribution and the specifics of the evaluation setup. These results underscore the promising potential of proposed method to advance the stateof-the-art in ECG-based disease detection, offering a valuable contribution to the field.

In contrast to the methods listed in Table 3, there exist alternative approaches, such as those referenced as [16], [17], [18], which do not involve the segmentation of ECG reports into individual lead images but instead process them holistically. However, it's important to note that these holistic methods operate on a different paradigm and problem scope compared to our proposed method. Consequently, they have been excluded from the comparative analysis to maintain consistency within the scope of this study.

## V. Conclusion

In this study, a comprehensive exploration of disease detection in ECG waveform images through the utilization of advanced vision transformer models has been undertaken. Specifically, three distinct vision transformer models, namely Google-Vit, Swin, and BEiT, have been employed. These models, designed to process lead images, have demonstrated remarkable performance, showcasing their potential in the realm of healthcare applications. Notably, BEiT has emerged as the frontrunner, with exceptional precision, recall, F1-score, and accuracy, positioning it as a compelling choice for accurate and reliable disease detection.

The comparative analysis, extending to benchmark models from recent literature, has highlighted the superiority of the proposed BEiT method. Across various evaluation scenarios, the method has consistently outperformed the benchmark models, underscoring its potential to advance the state-of-theart in ECG-based disease detection.

## References

[1] A. Rath, D. Mishra, G. Panda, and S. C. Satapathy, "Heart disease detection using deep learning methods from imbalanced ecg samples," *Biomedical Signal Processing and Control*, vol. 68, p. 102820, 2021.

[2] A. H. Khan, M. Hussain, and M. K. Malik, "Cardiac disorder classification by electrocardiogram sensing using deep neural network," *Complexity*, vol. 2021, pp. 1–8, 2021.

[3] M. S. Haleem, R. Castaldo, S. M. Pagliara, M. Petretta, M. Salvatore, M. Franzese, and L. Pecchia, "Time adaptive ecg driven cardiovascular disease detector," *Biomedical Signal Processing and Control*, vol. 70, p. 102968, 2021.

[4] S. Arooj, S. u. Rehman, A. Imran, A. Almuhaimeed, A. K. Alzahrani, and A. Alzahrani, "A deep convolutional neural network for the early detection of heart disease," *Biomedicines*, vol. 10, no. 11, p. 2796, 2022.

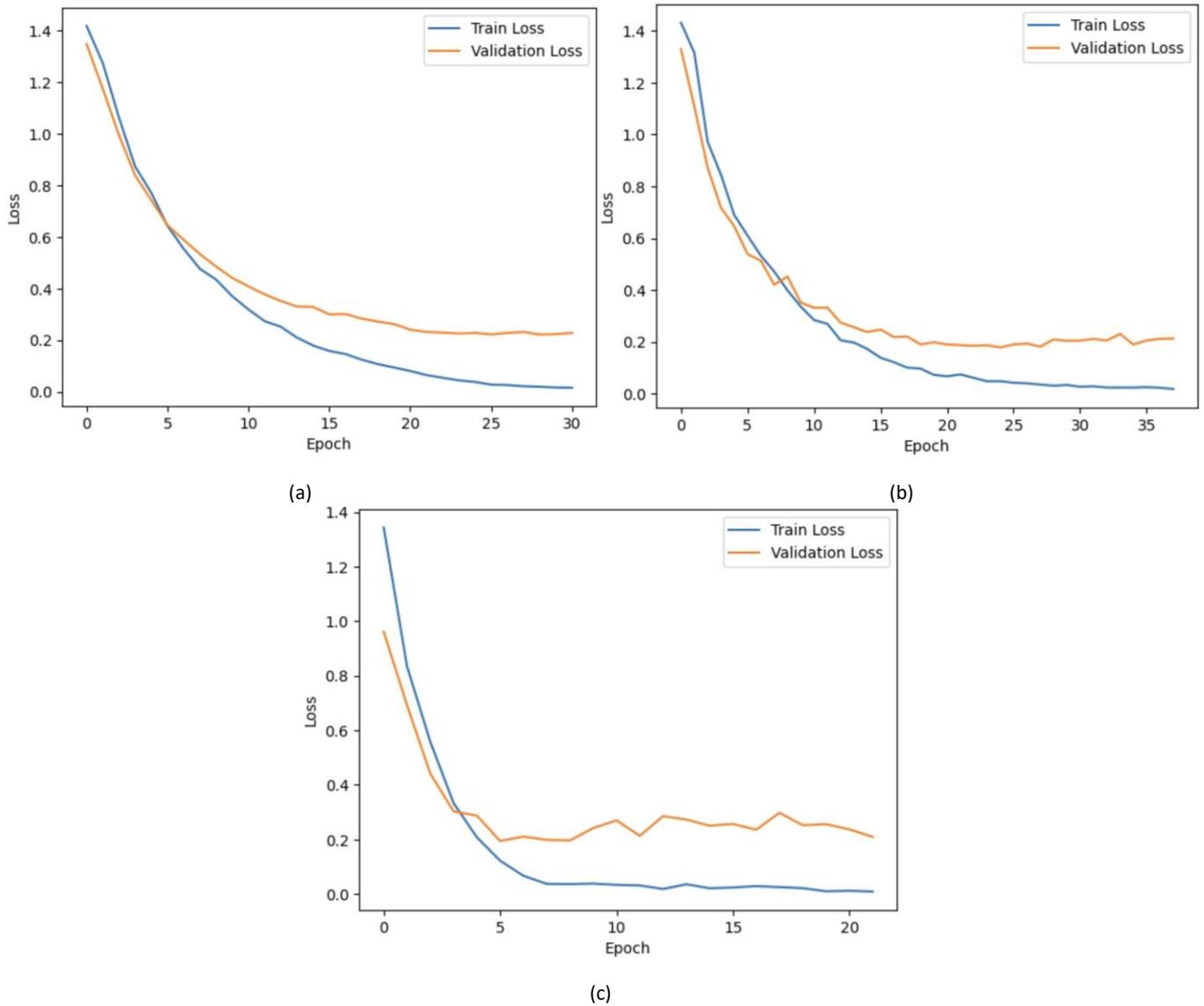

Fig. 6: Training loss curves (a) Google-Vit (b) Swin (c) BEiT

TABLE III: Comparison of proposed method with literature

| Models | Test Setup | Precision | Recall | F1-Score | Accuracy |
|---|---|---|---|---|---|
| [19] | 10-fold | N/A | 0.920 | 0.920 | 0.920 |
| [20] | 5-fold | N/A | N/A | N/A | 0.774 |
| [15] | Holdout 70/20/10 | N/A | 0.978 | 0.952 | 0.961 |
| [21] | N/A | 0.940 | 0.920 | 0.940 | N/A |
| Proposed Method | 5-Fold | 0.959 | 0.959 | 0.959 | 0.959 |
| Proposed Method | Holdout 80/0/20 | 0.966 | 0.966 | 0.966 | 0.966 |

[5] M. A. A. Ferreira, M. V. Gurgel, L. B. Marinho, N. M. M. Nascimento, S. P. P. da Silva, S. S. A. Alves, G. L. B. Ramalho, and P. P. Rebouças Filho, "Evaluation of heart disease diagnosis approach using ecg images," in *2019 International Joint Conference on Neural Networks (IJCNN)*. IEEE, 2019, pp. 1–7.

[6] H. Kim, M. I. M. Ishag, M. Piao, T. Kwon, and K. H. Ryu, "A data mining approach for cardiovascular disease diagnosis using heart rate variability and images of carotid arteries," *Symmetry*, vol. 8, no. 6, p. 47, 2016.

[7] M. Alsaffar, A. Alshammari, G. Alshammari, S. Aljaloud, T. S. Almurayziq, F. M. Abdoon, S. Abebaw *et al.*, "Machine learning for ischemic heart disease diagnosis aided by evolutionary computing," *Applied Bionics and Biomechanics*, vol. 2021, 2021.

[8] A. S. Eltrass, M. B. Tayel, and A. I. Ammar, "A new automated cnn deep learning approach for identification of ecg congestive heart failure and arrhythmia using constant-q non-stationary gabor transform," *Biomedical signal processing and control*, vol. 65, p. 102326, 2021.

[9] L. Mhamdi, O. Dammak, F. Cottin, and I. B. Dhaou, "Artificial intelligence for cardiac diseases diagnosis and prediction using ecg images on embedded systems," *Biomedicines*, vol. 10, no. 8, p. 2013, 2022.

[10] T. P. Naidu, K. A. Gopal, S. R. Ahmed, R. Revathi, S. H. Ahammad, V. Rajesh, S. Inthiyaz, and K. Saikumar, "A hybridized model for the prediction of heart disease using ml algorithms," in *2021 3rd International Conference on Advances in Computing, Communication Control and Networking (ICAC3N)*. IEEE, 2021, pp. 256–261.

[11] A. H. Khan, M. Hussain, and M. K. Malik, "Ecg images dataset of cardiac and covid-19 patients," *Data in Brief*, vol. 34, p. 106762, 2021. [12] A. Dosovitskiy, L. Beyer, A. Kolesnikov, D. Weissenborn, X. Zhai, T. Unterthiner, M. Dehghani, M. Minderer, G. Heigold, S. Gelly, J. Uszkoreit, and N. Houlsby, "An image is worth 16x16 words: Transformers for image recognition at scale," 2021.

[13] Z. Liu, Y. Lin, Y. Cao, H. Hu, Y. Wei, Z. Zhang, S. Lin, and B. Guo, "Swin transformer: Hierarchical vision transformer using shifted windows," *ICCV*, 2021.

[14] H. Bao, L. Dong, S. Piao, and F. Wei, "Beit: Bert pre-training of image transformers," in *ICLR*, 2022.

[15] J. Nagaraj and A. Leema, "Light weight multi-branch network-based extraction and classification of myocardial infarction from 12 lead electrocardiogram images," *The Imaging Science Journal*, vol. 71, no. 2, pp. 188–198, 2023.

[16] O. Attallah, "Ecg-biconet: An ecg-based pipeline for covid-19 diagnosis using bi-layers of deep features integration," *Computers in biology and medicine*, vol. 142, p. 105210, 2022.

[17] K. Prashant, P. Choudhary, T. Agrawal, and E. Kaushik, "Owae-net: Covid-19 detection from ecg images using deep learning and optimized weighted average ensemble technique," *Intelligent Systems with Applications*, vol. 16, p. 200154, 2022.

[18] N. Jothiaruna and A. A. Leema, "Lw-dn161: a cardiovascular disorder classification from 12 lead ecg images using convolutional neural network," *Soft Computing*, pp. 1–12, 2023.

[19] N. Sobahi, A. Sengur, R.-S. Tan, and U. R. Acharya, "Attentionbased 3d cnn with residual connections for efficient ecg-based covid-19 detection," *Computers in Biology and Medicine*, vol. 143, p. 105335, 2022.

[20] M. K. Chaitanya, L. D. Sharma, J. Rahul, D. Sharma, and A. Roy, "Artificial intelligence based approach for categorization of covid-19 ecg images in presence of other cardiovascular disorders," *Biomedical Physics & Engineering Express*, vol. 9, no. 3, p. 035012, 2023.

[21] V. Gururaj, S. P. Shankar, A. Bharadwaj *et al.*, "Electrocardiogram based cardiovascular disease detection with ensemble learning classifier," in *2022 4th International Conference on Circuits, Control, Communication and Computing (I4C)*, 2022, pp. 48–53.